\documentclass[conference]{IEEEtran}
\IEEEoverridecommandlockouts

\usepackage{cite}
\usepackage{amsmath,amssymb,amsfonts}
\usepackage{algorithm}
\usepackage{algorithmic}
\usepackage{graphicx}
\usepackage{textcomp}
\usepackage{xcolor}
\usepackage{booktabs}  
\usepackage{multirow}  
\usepackage{url}
\def\BibTeX{{\rm B\kern-.05em{\sc i\kern-.025em b}\kern-.08em
    T\kern-.1667em\lower.7ex\hbox{E}\kern-.125emX}}
\begin{document}

\title{Evidence Graph Consistency in Retrieval-Augmented Generation:
A Model-Dependent Analysis of Hallucination Detection}

\author{\IEEEauthorblockN{Jianru Shen}
\IEEEauthorblockA{\textit{University of Montana}\\
Missoula, MT, USA \\
js258133@umconnect.umt.edu \\
ORCID: 0009-0000-3546-9616
}}

\maketitle

\begin{abstract}
Retrieval-Augmented Generation (RAG) reduces but does not
eliminate hallucination in large language models. Existing
detection methods rely on flat similarity between generated
answers and retrieved passages, ignoring structural
relationships among evidence pieces and answer claims. We
propose Evidence Graph Consistency (EGC), a framework that
constructs a local evidence graph per response and computes
five structural consistency measures as hallucination
indicators. Evaluated on the full question answering split
of RAGTruth across six LLMs (5,767 responses), EGC reveals
a consistent model-family split: graph consistency features show the expected diagnostic
direction for hallucinations in Llama-2 models but
exhibit systematic reversal in GPT-4, GPT-3.5, and
Mistral-7B. This reversal suggests qualitatively different hallucination
patterns across model families and indicates that
embedding-based graph consistency cannot serve as a
model-independent hallucination detection signal.
\end{abstract}

\begin{IEEEkeywords}
retrieval-augmented generation, hallucination detection, graph
consistency, large language models, evidence graph
\end{IEEEkeywords}

\section{Introduction}

Retrieval-Augmented Generation (RAG) has become a standard
technique for grounding large language model (LLM) outputs
in external knowledge \cite{lewis2020rag, gao2024ragsurvey}. By conditioning
generation on retrieved passages, RAG reduces the frequency
of factually unsupported outputs \cite{shuster2021rag}
compared to parametric generation alone. Nevertheless, RAG does not eliminate hallucination: LLMs
continue to produce claims that contradict or extend beyond
the retrieved evidence \cite{wu2023ragtruth, ji2023survey,
huang2023survey}.

Existing approaches to hallucination detection in RAG
settings largely rely on flat similarity between the
generated answer and the retrieved context \cite{es2023ragas},
or on prompting a separate LLM to judge faithfulness
\cite{min2023factscore}. Both approaches treat the
relationship between answer and evidence as a single scalar
signal, ignoring the structural relationships among
individual evidence passages and answer claims.

We propose Evidence Graph Consistency (EGC), a lightweight
framework that constructs a local graph over the question,
retrieved passages, and answer claims, then computes
structural consistency features as hallucination indicators.
Applying EGC to the full question answering split of
RAGTruth \cite{wu2023ragtruth} across six LLMs, we find
that its effectiveness is model-dependent: EGC shows the expected diagnostic direction for Llama-2 models but exhibits systematic reversal for GPT-4, GPT-3.5, and
Mistral-7B, indicating qualitatively different hallucination
patterns across model families.

The contributions of this paper are as follows. We use EGC
as a structural probe to investigate when graph-based
consistency is and is not a valid hallucination signal
across model families, rather than to propose a competitive
detector. We conduct a systematic evaluation across six
LLMs on 5,767 RAG responses and reveal a consistent
model-family split with implications for the design of
hallucination detection systems. We analyse the structural patterns underlying this split
and demonstrate that embedding-based graph consistency alone
is not reliable as a model-independent hallucination signal.

\section{Related Work}

\textbf{Hallucination detection in RAG.}
RAGAs \cite{es2023ragas} decomposes generated answers into
atomic claims and checks each claim against the retrieved
context using an LLM judge, computing a faithfulness score
as the fraction of supported claims. ARES \cite{saad2023ares}
extends this paradigm by training lightweight classifiers
to evaluate RAG systems without requiring human annotations.
FActScore \cite{min2023factscore} applies a similar
decomposition strategy to long-form generation, verifying
claims against a retrieval corpus. SelfCheckGPT
\cite{manakul2023selfcheck} detects hallucinations by
sampling multiple responses and measuring self-consistency,
without requiring external evidence. LRP4RAG \cite{hu2024lrp4rag} detects hallucinations via
layer-wise relevance propagation over the model's internal
states, providing an interpretable signal that does not
require external annotations. RAG-HAT \cite{song2024raghat}
takes a complementary approach by fine-tuning LLMs with
hallucination-aware preference data to reduce hallucination
at generation time. Unlike these approaches, EGC operates
on graph topology derived from embedding similarity rather
than LLM-generated judgements or stochastic sampling,
making it computationally lightweight and model-agnostic.

\textbf{Graph-based approaches in NLP.}
Graph structures have been used for multi-hop question
answering \cite{yang2018hotpotqa,yasunaga2021qagnn},
knowledge-grounded dialogue \cite{dinan2019wizard}, document
summarisation \cite{maynez2020faithfulness,falke2019ranking},
and factuality assessment
\cite{ribeiro2022factgraph,sansford2024grapheval}. These
approaches typically rely on large pre-constructed knowledge
graphs. EGC instead constructs a lightweight local graph per
response at inference time, requiring no external graph
infrastructure or pre-processing.

\textbf{The RAGTruth corpus.}
RAGTruth \cite{wu2023ragtruth} provides 17,790 naturally
generated responses from six LLMs across QA, data-to-text,
and summarisation tasks, with word-level hallucination
annotations produced by human labellers. It is a large-scale RAG hallucination corpus with
span-level annotations across multiple model families
\cite{fan2024ragsurvey}, making it suitable for the
cross-model analysis we conduct.

\textbf{Model-dependent evaluation.}
Prior work has noted that hallucination rates vary
substantially across model families
\cite{wu2023ragtruth, ji2023survey, huang2023survey,
touvron2023llama2, jiang2023mistral, openai2023gpt4}, but the downstream effect on detection
method effectiveness has not been systematically studied. Our results show that this variation is qualitative: the
diagnostic direction of EGC measures differs across the
evaluated model families. Unlike prior work that aims to maximise
detection accuracy, this paper uses EGC as a diagnostic
lens to examine when structural consistency is and is not
a valid hallucination signal.

\section{Method}
\label{sec:method}

We propose Evidence Graph Consistency (EGC), a lightweight
framework that constructs a local evidence graph per RAG
response and derives structural consistency features as
hallucination indicators.

\subsection{Graph Construction}

Given a question $q$, a set of retrieved passages
$\mathcal{P} = \{p_1, \ldots, p_k\}$, and a generated answer $a$, we
construct an undirected graph $G = (V, \mathcal{E})$ with three node types.

\textbf{Nodes.} A single \textit{question node} $v_q$ represents the
input question. Each passage $p_i$ becomes an \textit{evidence node}
$v_{e_i}$. The answer $a$ is segmented into sentences using spaCy
\cite{spacy}, and each sentence of length greater than ten tokens becomes
a \textit{claim node} $v_{c_j}$. All nodes are encoded with
\texttt{all-MiniLM-L6-v2} \cite{reimers2019sentencebert, devlin2019bert}.
Throughout, $\mathcal{E}$ denotes the edge set of $G$, while $V_E$
and $V_C$ denote the sets of evidence and claim nodes
respectively.

\textbf{Edges.} We define three edge types based on cosine similarity
$\text{sim}(\cdot,\cdot)$ with threshold $\tau = 0.4$:

Cosine similarity is chosen because all nodes are encoded
as fixed-dimensional dense vectors by \texttt{all-MiniLM-L6-v2};
the similarity between any two nodes is therefore
independent of the original text length and is defined as
$\text{sim}(\mathbf{u}, \mathbf{v}) = \frac{\mathbf{u} \cdot
\mathbf{v}}{\|\mathbf{u}\| \|\mathbf{v}\|}$, where
$\mathbf{u}, \mathbf{v} \in \mathbb{R}^{384}$ are the
corresponding embeddings \cite{mikolov2013word2vec}.

\begin{itemize}
  \item \textit{Q--E edges}: connect $v_q$ to $v_{e_i}$ if
        $\text{sim}(v_q, v_{e_i}) \geq \tau$
  \item \textit{E--C edges}: connect $v_{e_i}$ to $v_{c_j}$ if
        $\text{sim}(v_{e_i}, v_{c_j}) \geq \tau$
  \item \textit{E--E edges}: connect $v_{e_i}$ to $v_{e_l}$ if
        $\text{sim}(v_{e_i}, v_{e_l}) \geq \tau$ and $i \neq l$
\end{itemize}

\begin{figure*}[htbp]
\centerline{\includegraphics[width=\textwidth]{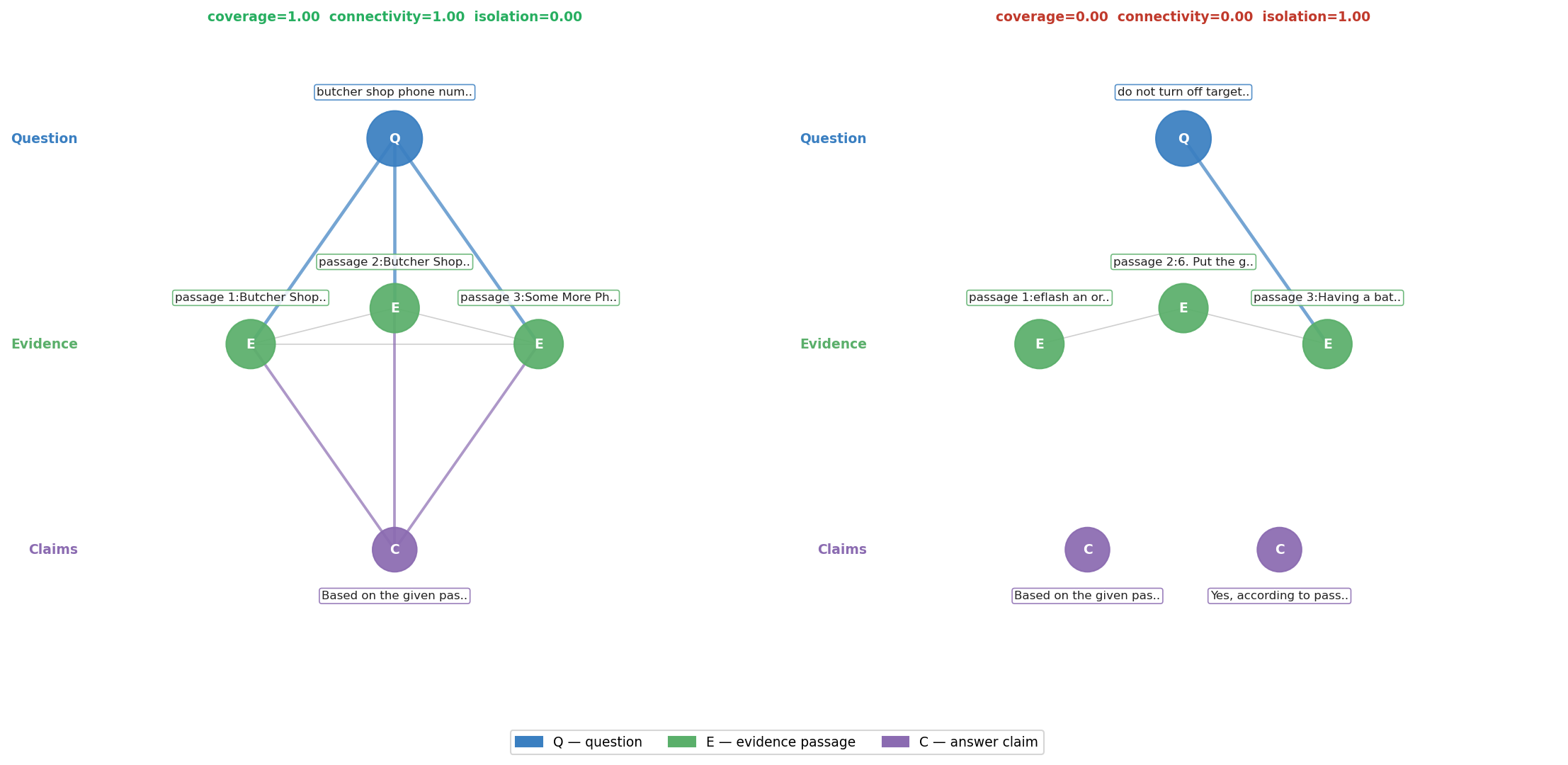}}
\caption{Evidence graph structure for a grounded answer (left) and a
hallucinated answer (right) from Llama-2-13B. In the grounded case all
claim nodes connect to evidence; in the hallucinated case claim nodes
are fully isolated.}
\label{fig:graph}
\end{figure*}

Fig.~\ref{fig:graph} illustrates the resulting graph structure for a
grounded and a hallucinated answer.

\subsection{Consistency Features}

We compute five scalar measures from $G$, where $d(v)$
denotes the degree of node $v$.

\textbf{Coverage} measures the fraction of claims supported by at
least one evidence node:
\begin{equation}
  \text{cov} = \frac{|\{v_c \in V_C : \exists\, v_e \in
  V_E,\, (v_c, v_e) \in \mathcal{E}\}|}{|V_C|}
\end{equation}

\textbf{Support density} measures the average number of evidence nodes
per claim, normalised by the total number of evidence nodes:
\begin{equation}
  \text{sup} = \frac{1}{|V_C|} \sum_{v_c \in V_C}
  \frac{|\{v_e \in V_E : (v_c, v_e) \in \mathcal{E}\}|}{|V_E|}
\end{equation}

\textbf{Cross-evidence agreement} is the mean weight of all E--E edges,
capturing whether retrieved passages are mutually consistent:
\begin{equation}
  \text{agr} = \frac{\sum_{(v_e, v_{e'}) \in \mathcal{E}_{\text{EE}}}
  \text{sim}(v_e, v_{e'})}{|\mathcal{E}_{\text{EE}}|}
\end{equation}
where $\mathcal{E}_{\text{EE}}$ is the set of E--E edges. If no E--E edges exist,
$\text{agr} = 0$.

\textbf{Connectivity} measures the fraction of claim nodes reachable
from the question node via any path in $G$:
\begin{equation}
  \text{conn} = \frac{|\{v_c \in V_C :
  v_q \leadsto v_c \text{ in } G\}|}{|V_C|}
\end{equation}

\textbf{Isolation penalty} is the fraction of claim nodes with no
edges, indicating claims entirely unsupported by retrieved evidence:
\begin{equation}
  \text{iso} = \frac{|\{v_c \in V_C : d(v_c) = 0\}|}
  {|V_C|}
\end{equation}

Because claim nodes connect only to evidence nodes, the
isolation penalty is the complement of coverage, that is,
$\mathrm{iso} = 1 - \mathrm{cov}$. We retain both quantities
because they offer complementary interpretations: coverage
emphasises supported claims, whereas isolation directly
highlights unsupported ones. They are not independent
structural signals.

\begin{algorithm}
\caption{EGC Graph Construction}
\label{alg:graph}
\begin{algorithmic}[1]
\REQUIRE Question $q$, passages $\mathcal{P} = \{p_1,\ldots,p_k\}$,
         answer $a$, similarity threshold $\tau$
\ENSURE Evidence graph $G = (V, \mathcal{E})$
\STATE Encode $q$, each $p_i \in \mathcal{P}$, and each sentence
       $c_j$ of $a$ with \texttt{all-MiniLM-L6-v2} to obtain
       embeddings $\mathbf{e}_q$, $\{\mathbf{e}_{p_i}\}$,
       $\{\mathbf{e}_{c_j}\}$
\STATE $V \leftarrow \{v_q\} \cup \{v_{e_i}\} \cup \{v_{c_j}\}$
\STATE $\mathcal{E} \leftarrow \emptyset$
\FOR{each $v_{e_i} \in V_E$}
    \IF{$\text{sim}(\mathbf{e}_q, \mathbf{e}_{p_i}) \geq \tau$}
        \STATE $\mathcal{E} \leftarrow \mathcal{E} \cup \{(v_q, v_{e_i})\}$
    \ENDIF
\ENDFOR
\FOR{each $v_{e_i} \in V_E$, each $v_{c_j} \in V_C$}
    \IF{$\text{sim}(\mathbf{e}_{p_i}, \mathbf{e}_{c_j}) \geq \tau$}
        \STATE $\mathcal{E} \leftarrow \mathcal{E} \cup \{(v_{e_i}, v_{c_j})\}$
    \ENDIF
\ENDFOR
\FOR{each pair $v_{e_i}, v_{e_l} \in V_E$, $i \neq l$}
    \IF{$\text{sim}(\mathbf{e}_{p_i}, \mathbf{e}_{p_l}) \geq \tau$}
        \STATE $\mathcal{E} \leftarrow \mathcal{E} \cup \{(v_{e_i}, v_{e_l})\}$
    \ENDIF
\ENDFOR
\RETURN $G = (V, \mathcal{E})$
\end{algorithmic}
\end{algorithm}

\begin{algorithm}
\caption{EGC Feature Computation}
\label{alg:features}
\begin{algorithmic}[1]
\REQUIRE Evidence graph $G = (V, \mathcal{E})$, node sets
         $V_E$, $V_C$, question node $v_q$
\ENSURE Feature vector
        $\mathbf{f} = [\text{cov}, \text{sup}, \text{agr},
        \text{conn}, \text{iso}]$

\STATE $n_c \leftarrow |V_C|$

\STATE $\text{cov} \leftarrow
       \frac{1}{n_c} \sum_{v_c \in V_C}
       \mathbf{1}[\exists\, v_e \in V_E :
       (v_c, v_e) \in \mathcal{E}]$

\STATE $\text{sup} \leftarrow
       \frac{1}{n_c \cdot |V_E|}
       \sum_{v_c \in V_C}
       |\{v_e \in V_E : (v_c, v_e) \in \mathcal{E}\}|$

\STATE $\mathcal{E}_{\text{EE}} \leftarrow
       \{(v_e, v_{e'}) \in \mathcal{E} : v_e, v_{e'} \in V_E\}$
\IF{$|\mathcal{E}_{\text{EE}}| > 0$}
    \STATE $\text{agr} \leftarrow
           \frac{1}{|\mathcal{E}_{\text{EE}}|}
           \sum_{(v_e,v_{e'}) \in \mathcal{E}_{\text{EE}}}
           \text{sim}(\mathbf{e}_{v_e}, \mathbf{e}_{v_{e'}})$
\ELSE
    \STATE $\text{agr} \leftarrow 0$
\ENDIF

\STATE $\text{conn} \leftarrow
       \frac{1}{n_c} \sum_{v_c \in V_C}
       \mathbf{1}[v_q \leadsto v_c \text{ in } G]$

\STATE $\text{iso} \leftarrow
       \frac{1}{n_c} \sum_{v_c \in V_C}
       \mathbf{1}[d(v_c) = 0]$

\RETURN $\mathbf{f} = [\text{cov}, \text{sup}, \text{agr},
        \text{conn}, \text{iso}]$
\end{algorithmic}
\end{algorithm}

\subsection{Hallucination Diagnosis}

The five measures $\mathbf{f} = [\text{cov}, \text{sup},
\text{agr}, \text{conn}, \text{iso}]$ are standardised and
passed to a logistic regression classifier
\cite{pedregosa2011sklearn, hosmer2013logistic} with
balanced class weights.
We use the official train and test splits of RAGTruth
\cite{wu2023ragtruth} and report AUROC, F1, and the
per-model EGC diagnostic gap:
\begin{equation}
  \Delta_m = \overline{\text{EGC}}_{\text{grounded}}^{(m)} -
  \overline{\text{EGC}}_{\text{hallucinated}}^{(m)}
\end{equation}
where $\overline{\text{EGC}}^{(m)}$ is the mean EGC score
for model $m$ and the composite EGC score is defined as:
\begin{equation}
  \text{EGC}(a) = \frac{\text{cov} + \text{sup} +
  \text{conn} - \text{iso}}{3}
\end{equation}

The score combines coverage, support density, and connectivity
as positive indicators of evidence grounding, and subtracts
isolation penalty as a negative indicator of unsupported
claims. Because $\mathrm{iso} = 1 - \mathrm{cov}$, the
composite score places additional emphasis on whether
claims are supported by the retrieved evidence. The denominator normalises
the score to the range
$[-\frac{1}{3}, 1]$. Cross-evidence agreement is excluded
from the composite score because it measures inter-passage
consistency rather than claim-level grounding. Although it
provides limited aggregate diagnostic value,
Table~\ref{tab:perfeature} shows that it can capture
model-specific behaviour, including a strong reversed signal
for GPT-4.

A positive $\Delta_m$ indicates that EGC assigns higher scores to
grounded answers, consistent with the intended diagnostic direction.

\section{Experimental Setup}

\subsection{Dataset}

We evaluate on the question answering subset of RAGTruth
\cite{wu2023ragtruth}, a manually annotated hallucination corpus for
RAG applications. The QA subset contains 5,767 responses generated by
six LLMs across 989 unique questions drawn from MS MARCO \cite{msmarco}. RAGTruth
provides word-level hallucination annotations produced by human
labellers, categorising each hallucination span into one of four
types: evident conflict, subtle conflict, evident baseless information,
and subtle baseless information. We derive a binary response-level label by treating any
response containing at least one annotated span of any
type as hallucinated. This response-level label is
appropriate because EGC features are computed over the
full response graph rather than at the token or span
level; a binary label aligns the graph-level structural
signal with the response-level hallucination status.
We acknowledge that this binarisation does not exploit
the severity or type of individual hallucination spans;
future work could leverage span-level annotations for
claim-level or edge-level supervision. Under this definition, 1,706 of 5,767
responses (29.6\%) are hallucinated, yielding a class ratio of
approximately 1:2.4. We use the official train and test splits
provided by the dataset authors (4,892 train, 875 test).

Table~\ref{tab:datasplit} summarises the per-model sample distribution
across train and test splits. Hallucination rates vary substantially
across models, from 4.4\% for GPT-4 to 51.2\% for Llama-2-7B,
reflecting the well-documented relationship between model capability
and hallucination frequency \cite{wu2023ragtruth}.

\begin{table}[htbp]
\caption{Per-Model Sample Distribution in RAGTruth QA Split}
\label{tab:datasplit}
\begin{center}
\begin{tabular}{lcccc}
\toprule
\textbf{Model} & \textbf{Total} & \textbf{Train} & \textbf{Test}
    & \textbf{Hal.\%} \\
\midrule
GPT-4          & 961  & 815  & 146 &  4.4 \\
GPT-3.5        & 930  & 789  & 141 &  8.1 \\
Mistral-7B     & 939  & 797  & 142 & 40.3 \\
Llama-2-70B    & 982  & 834  & 148 & 32.4 \\
Llama-2-13B    & 984  & 835  & 149 & 40.2 \\
Llama-2-7B     & 971  & 822  & 149 & 51.2 \\
\midrule
Total          & 5767 & 4892 & 875 & 29.6 \\
\bottomrule
\multicolumn{5}{l}{\small Hal.\% = percentage of hallucinated
responses.}
\end{tabular}
\end{center}
\end{table}

\subsection{Models and Implementation}

All text representations are computed with
\texttt{all-MiniLM-L6-v2} \cite{reimers2019sentencebert}, a
22M-parameter sentence encoder trained on over one billion sentence
pairs. Sentence segmentation uses spaCy
\texttt{en\_core\_web\_sm} \cite{spacy}; sentences of ten tokens or
fewer are discarded to avoid short, potentially uninformative claim nodes.
The similarity threshold is set to $\tau = 0.4$
throughout all experiments. This fixed value is not
claimed to be universally optimal across all models or
domains; rather, a single threshold is used deliberately
so that any observed differences in diagnostic direction
across model families can be attributed to the models
themselves rather than to threshold variation. Graph construction and feature computation are implemented
with NetworkX \cite{networkx}.

The classifier is logistic regression \cite{pedregosa2011sklearn} with
\texttt{class\_weight=balanced}, which reweights each training sample
by the inverse frequency of its class. This addresses the 1:2.4 class
imbalance without requiring oversampling or synthetic data generation.
Features are standardised to zero mean and unit variance before
classification.
Because this study uses the released RAGTruth responses,
no additional generation prompts are introduced in our
experiments. Given the fixed dataset, sentence segmenter,
encoder, and similarity threshold, graph construction is
fully deterministic. This makes the reported EGC features
reproducible without requiring access to the original LLM
inference endpoints. Code is available at \url{https://doi.org/10.5281/zenodo.20932843}. The only learned component is the
logistic regression classifier, which is trained on the
official RAGTruth training split and evaluated on the
official test split.

\subsection{Evaluation}

We report three metrics. Area under the ROC curve (AUROC) is the
primary metric as it is threshold-independent and measures
discrimination ability across the full range of classification
thresholds, making it appropriate for imbalanced binary classification.
Macro F1 \cite{vanrijsbergen1979}, the unweighted mean
of per-class F1 scores, is reported as a secondary
metric; it treats each class equally and is appropriate
for imbalanced classification where both precision and
recall matter. The per-model diagnostic gap $\Delta_m$, defined in
Section~\ref{sec:method}, quantifies the direction and magnitude of
EGC effectiveness for each model family. All metrics are computed on
the official test split only; the train split is used exclusively for
fitting the logistic regression classifier.

\begin{figure*}[htbp]
\centerline{\includegraphics[width=\textwidth]{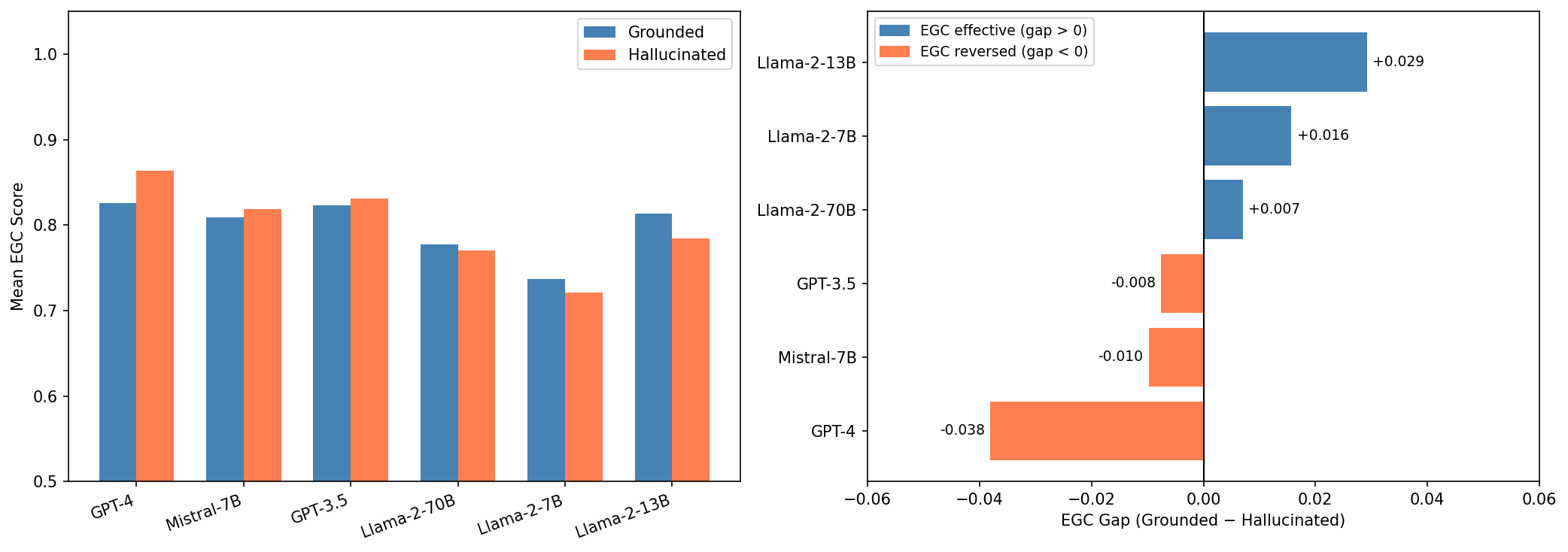}}
\caption{Left: mean EGC score by model and label. Right: per-model
diagnostic gap $\Delta_m$. Blue bars indicate models where EGC is
effective; orange bars indicate reversal.}
\label{fig:gap}
\end{figure*}

\begin{figure*}[htbp]
\centerline{\includegraphics[width=\textwidth]{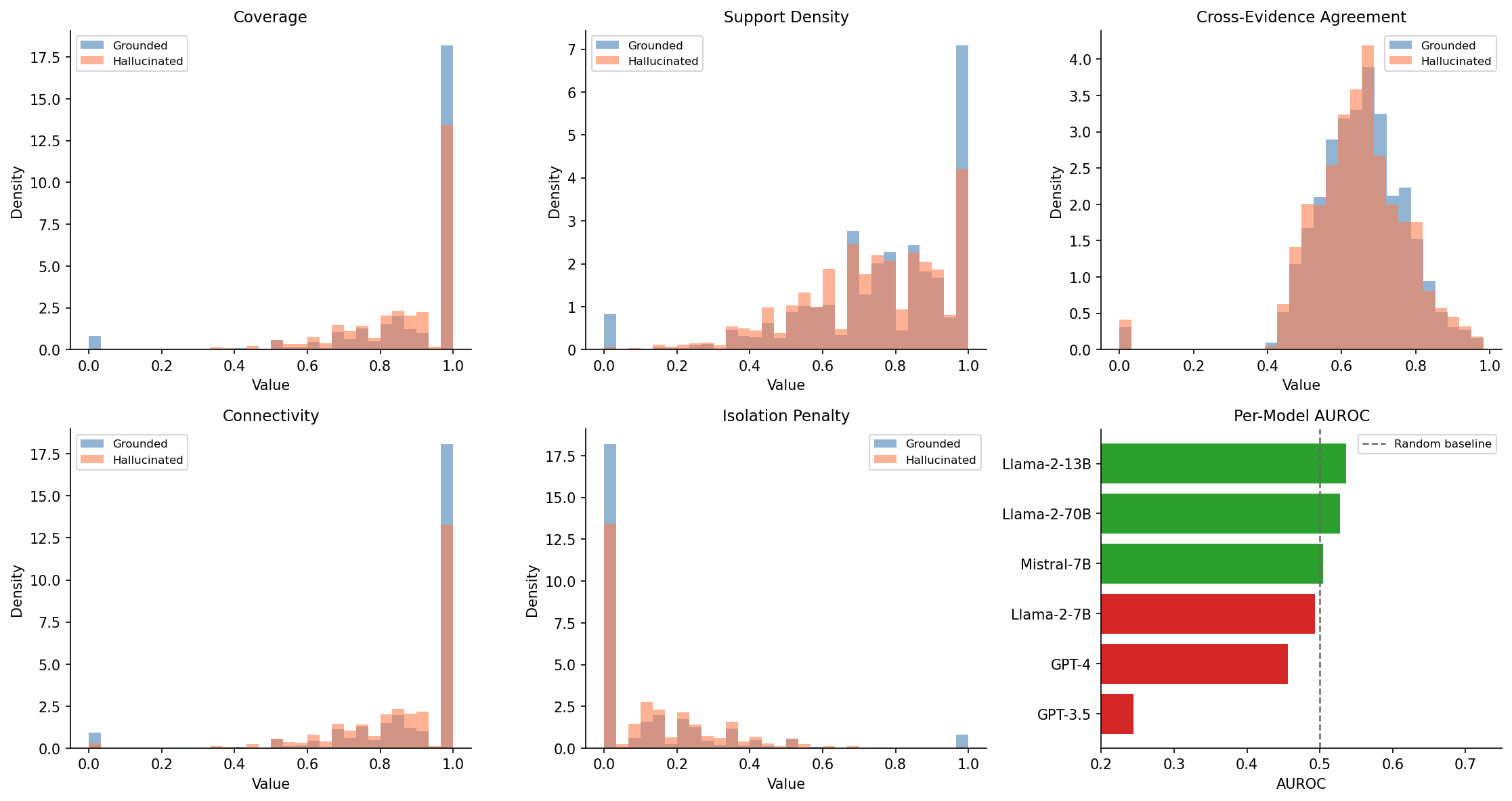}}
\caption{EGC feature distributions for grounded and hallucinated
answers across all models (top five panels), and per-model AUROC
(bottom right), where green and red bars denote AUROC above and
below the 0.5 random baseline respectively. GPT-3.5 AUROC of 0.244
reflects the systematic reversal described in Section~\ref{sec:permodel}.}
\label{fig:dist}
\end{figure*}

\section{Results}

\subsection{Overall Diagnostic Performance}

Table~\ref{tab:overall} reports AUROC and F1 on the full
test set. The coverage-only baseline achieves the strongest
aggregate AUROC of 0.589, while EGC reaches 0.556. This
suggests that aggregate pooling obscures the model-dependent
behaviour examined in the following sections.

Aggregate AUROC assumes a monotonic relationship between
EGC score and hallucination label across all samples. When that relationship flips sign between model families,
as our per-model analysis shows, pooling across families
cancels the signal rather than measuring it. The appropriate
diagnostic lens for this question is therefore the per-model
direction and magnitude of $\Delta_m$, which directly
captures whether structural graph consistency aligns with
the expected hallucination signal for each model family.

\begin{table}[htbp]
\caption{Overall Hallucination Detection Performance on RAGTruth QA Test Set}
\label{tab:overall}
\begin{center}
\begin{tabular}{lcc}
\toprule
\textbf{Method} & \textbf{AUROC} & \textbf{F1} \\
\midrule
Coverage only       & 0.589 & 0.296 \\
Support density only & 0.557 & 0.289 \\
EGC (all features)  & 0.556 & 0.285 \\
\bottomrule
\end{tabular}
\end{center}
\end{table}

Table~\ref{tab:family} further decomposes this result by
model family and applies a direction correction. We group Mistral-7B with GPT-class models because it
exhibits the same reversed diagnostic direction as confirmed by the per-model analysis in Section~\ref{sec:permodel}. When models are grouped by family, Llama models alone yield
AUROC 0.524 and GPT/Mistral models alone yield AUROC 0.509,
both modest. However, when the EGC score is sign-flipped for the
GPT/Mistral family before pooling, reflecting the reversal documented in Section~\ref{sec:permodel}, the
direction-corrected AUROC rises to 0.669, an improvement
of $+0.113$ over the uncorrected aggregate. This is not a performance claim but an interpretive one:
it demonstrates that the aggregate suppression is a direct
consequence of combining opposing signals, and that EGC
carries direction-heterogeneous diagnostic information
across model families rather than a uniformly weak signal.
The pooled metric therefore underestimates the diagnostic
value of a signal that is consistent within families but
opposed between them.

\begin{table}[htbp]
\caption{Family-Level and Direction-Corrected AUROC}
\label{tab:family}
\begin{center}
\begin{tabular}{lcc}
\toprule
\textbf{Evaluation} & \textbf{AUROC} & \textbf{$n$} \\
\midrule
Llama family only        & 0.524 & 448 \\
GPT + Mistral only       & 0.509 & 427 \\
\midrule
All models (overall)     & 0.556 & 875 \\
All models (direction-corrected) & \textbf{0.669} & 875 \\
\bottomrule
\multicolumn{3}{l}{\small Direction-corrected: EGC scores
sign-flipped for GPT/Mistral} \\
\multicolumn{3}{l}{\small before pooling, reflecting the
observed family-level reversal.}
\end{tabular}
\end{center}
\end{table}

\subsection{Per-Model Analysis}
\label{sec:permodel}

Table~\ref{tab:permodel} breaks down AUROC and the diagnostic gap
$\Delta_m$ by model. A clear split emerges between the Llama family
and the remaining models.

\begin{table}[htbp]
\caption{Per-Model EGC Diagnostic Gap and AUROC}
\label{tab:permodel}
\begin{center}
\begin{tabular}{lcccc}
\toprule
\textbf{Model} & \textbf{Hal.\%} & \textbf{AUROC}
    & $\boldsymbol{\Delta_m}$ & \textbf{Direction} \\
\midrule
Llama-2-13B  & 40.2 & 0.536 & $+$0.029 & Correct   \\
Llama-2-7B   & 51.2 & 0.493 & $+$0.016 & Correct   \\
Llama-2-70B  & 32.4 & 0.528 & $+$0.007 & Correct   \\
\midrule
Mistral-7B   & 40.3 & 0.505 & $-$0.010 & Reversed  \\
GPT-3.5      &  8.1 & 0.244 & $-$0.008 & Reversed  \\
GPT-4        &  4.4 & 0.456 & $-$0.038 & Reversed  \\
\bottomrule
\multicolumn{5}{l}{\small Hal.\% = overall hallucination rate.} \\
\multicolumn{5}{l}{\small Direction is determined by the sign of
$\Delta_m$, not by whether} \\
\multicolumn{5}{l}{\small AUROC is above or below 0.5.}
\end{tabular}
\end{center}
\end{table}

For all three Llama models, $\Delta_m > 0$: grounded answers receive
higher EGC scores than hallucinated ones, consistent with the intended
diagnostic direction. GPT-4 shows the strongest reversal at
$\Delta_m = -0.038$, meaning its hallucinated answers score
systematically higher than its grounded ones. GPT-3.5 and Mistral-7B
exhibit smaller but consistent reversals.

Fig.~\ref{fig:gap} visualises the EGC score distributions and
per-model gap, showing that Llama hallucinations are
structurally sparser while GPT-4 hallucinations are not.

\subsection{Feature Distributions}
\label{sec:features}

Fig.~\ref{fig:dist} shows the distribution of all five EGC features
across the full test set. Coverage and connectivity are bimodal, with
a large mass at 1.0 for grounded answers and a heavier tail near 0 for
hallucinated ones. Cross-evidence agreement shows similar aggregate
distributions for both classes, although the per-model
analysis in Table~V reveals a strong reversed signal for
GPT-4. Isolation penalty mirrors the coverage distribution
as expected by construction.

\subsection{Case Study}

We examine two representative cases from Llama-2-13B to illustrate
when EGC succeeds and fails.

\textbf{Grounded answer} (EGC score 1.00). For the question
\textit{``butcher shop phone number''}, the model correctly extracts
a phone number from the retrieved passages. All three claim nodes
connect to at least one evidence node, yielding coverage = 1.00,
connectivity = 1.00, and isolation = 0.00.

\textbf{Hallucinated answer} (EGC score $-$0.33). For the question
\textit{``do not turn off target tizen''}, the model produces an
answer referencing a SIM card, a detail absent from the retrieved
passages. Both claim nodes are fully isolated with no evidence
connections, yielding coverage = 0.00, connectivity = 0.00, and
isolation = 1.00. Fig.~\ref{fig:graph} illustrates the structural
difference between these two cases.

For GPT-4, the reversal pattern is demonstrated by the question
\textit{``how to clean under laptop keyboard without removing keys''}.
The hallucinated answer scores EGC = 1.00 because its phrasing closely
mirrors the retrieved passages even though it introduces an unsupported
procedural claim. This linguistic alignment with the evidence corpus
causes all five features to signal grounded behaviour, showing that
EGC is unable to detect hallucinations that are semantically proximate
to the evidence.

\subsection{Robustness Analysis}
\label{sec:robust}

\textbf{Per-feature analysis.}
Table~\ref{tab:perfeature} reports per-feature AUROC for each
model on the test split. Coverage and connectivity show
consistent behaviour across Llama models, with AUROC modestly
above chance, while values fall below chance for GPT-3.5,
reflecting the reversal pattern observed in the aggregate EGC
score. Notably, cross-evidence agreement
shows an elevated AUROC of 0.772 for GPT-4, but in the
reversed direction: hallucinated GPT-4 answers exhibit higher
inter-passage agreement than grounded ones. Rather than
contradicting the failure mode interpretation, this reinforces
it. GPT-4 hallucinations closely mirror the retrieved
passages, producing dense and mutually consistent E--E
structure even when introducing unsupported claims.

\begin{table}[htbp]
\caption{Per-Feature AUROC by Model on Test Split}
\label{tab:perfeature}
\begin{center}
\begin{tabular}{lccccc}
\toprule
\textbf{Model} & \textbf{Cov.} & \textbf{Sup.}
    & \textbf{Agr.} & \textbf{Conn.} & \textbf{Iso.} \\
\midrule
Llama-2-13B  & 0.507 & 0.518 & 0.567 & 0.516 & 0.507 \\
Llama-2-7B   & 0.514 & 0.529 & 0.468 & 0.516 & 0.514 \\
Llama-2-70B  & 0.526 & 0.463 & 0.498 & 0.522 & 0.526 \\
\midrule
Mistral-7B   & 0.419 & 0.562 & 0.473 & 0.406 & 0.419 \\
GPT-3.5      & 0.293 & 0.365 & 0.515 & 0.278 & 0.293 \\
GPT-4        & 0.601 & 0.396 & 0.772 & 0.611 & 0.601 \\
\bottomrule
\end{tabular}

\vspace{2pt}
{\small
\begin{minipage}{\columnwidth}
\raggedright
Cov.=Coverage, Sup.=Support density, Agr.=Agreement,
Conn.=Connectivity, Iso.=Isolation penalty.
Coverage and isolation penalty are complementary; their
per-feature AUROC values coincide by construction.
\end{minipage}
}
\end{center}
\end{table}

\textbf{Threshold sensitivity.}
A potential concern is that the model-family split is an
artifact of the fixed threshold $\tau = 0.4$ rather than a
structural signal. Table~\ref{tab:threshold} reports
the diagnostic gap $\Delta_m$ across five thresholds from 0.3
to 0.7, along with the direction consistency score: the
fraction of models whose gap sign matches the expected
direction (positive for Llama, negative for GPT and Mistral).

At $\tau = 0.3$, graphs are overly dense because nearly all
node pairs exceed the low threshold, compressing structural
differences and yielding a consistency score of 0.833, with
Llama-2-70B showing an incorrect direction. At $\tau = 0.4$,
the model-family split is most pronounced: all six models
show the expected direction, yielding a perfect consistency
score of 1.000. At $\tau \geq 0.5$, graphs become
increasingly sparse as fewer edges meet the higher threshold.
Most claim nodes become isolated regardless of model family,
which causes hallucinated and grounded answers to converge
in EGC score and progressively degrades directional
consistency to 0.667 at $\tau = 0.5$ and 0.500 at
$\tau \geq 0.6$.

These results confirm two things. First, $\tau = 0.4$ is
empirically the most discriminative operating point in our
study: it is the unique threshold at which all six models
show the expected directional split simultaneously. This
supports that the model-family reversal is unlikely to be
a trivial artifact of threshold choice. Second, the
degradation pattern is structurally predictable: at lower
thresholds graphs become too dense to preserve
discriminative structure, while at higher thresholds graphs
become too sparse and all models converge toward high
isolation regardless of label. The fact that consistency
degrades in opposite directions for opposite reasons further
supports $\tau = 0.4$ as a meaningful operating point rather
than an arbitrary one.

\begin{table*}[htbp]
\caption{Diagnostic Gap $\Delta_m$ and Direction Consistency
Across Similarity Thresholds}
\label{tab:threshold}
\begin{center}
\begin{tabular}{lccccccc}
\toprule
$\boldsymbol{\tau}$ & \textbf{Llama-2-13B} & \textbf{Llama-2-7B}
    & \textbf{Llama-2-70B} & \textbf{Mistral-7B}
    & \textbf{GPT-3.5} & \textbf{GPT-4}
    & \textbf{Score} \\
\midrule
0.3 & $+$0.008 & $+$0.002 & $-$0.022 & $-$0.036
    & $-$0.046 & $-$0.057 & 0.833 \\
\textbf{0.4} & $\mathbf{+0.029}$ & $\mathbf{+0.016}$
    & $\mathbf{+0.007}$ & $\mathbf{-0.010}$
    & $\mathbf{-0.008}$ & $\mathbf{-0.038}$
    & \textbf{1.000} \\
0.5 & $+$0.074 & $+$0.031 & $+$0.027 & $+$0.025
    & $+$0.048 & $-$0.023 & 0.667 \\
0.6 & $+$0.092 & $+$0.037 & $+$0.041 & $+$0.094
    & $+$0.102 & $+$0.019 & 0.500 \\
0.7 & $+$0.115 & $+$0.035 & $+$0.052 & $+$0.113
    & $+$0.085 & $+$0.060 & 0.500 \\
\bottomrule
\multicolumn{8}{l}{\small Score = fraction of models with
correct direction. $\tau=0.4$ (bold) achieves perfect
consistency across all six models.}
\end{tabular}
\end{center}
\end{table*}

Table~\ref{tab:direction} shows the per-model directional
correctness at each threshold. At $\tau = 0.4$, all six
models show the correct direction. At $\tau = 0.3$,
Llama-2-70B is the only exception. At $\tau \geq 0.5$, GPT-3.5 and GPT-4 shift to incorrect
directions; Mistral-7B follows at $\tau \geq 0.6$.

\begin{table}[htbp]
\caption{Per-Model Directional Correctness Across Thresholds}
\label{tab:direction}
\begin{center}
\begin{tabular}{lcccccc}
\toprule
$\boldsymbol{\tau}$ & \textbf{L-13B} & \textbf{L-7B}
    & \textbf{L-70B} & \textbf{Mis.}
    & \textbf{G-3.5} & \textbf{G-4} \\
\midrule
0.3 & \checkmark & \checkmark & \texttimes
    & \checkmark & \checkmark & \checkmark \\
\textbf{0.4} & \checkmark & \checkmark & \checkmark
    & \checkmark & \checkmark & \checkmark \\
0.5 & \checkmark & \checkmark & \checkmark
    & \checkmark & \texttimes & \texttimes \\
0.6 & \checkmark & \checkmark & \checkmark
    & \texttimes & \texttimes & \texttimes \\
0.7 & \checkmark & \checkmark & \checkmark
    & \texttimes & \texttimes & \texttimes \\
\bottomrule
\multicolumn{7}{l}{\small \checkmark = correct direction,
\texttimes = incorrect.} \\
\multicolumn{7}{l}{\small L=Llama-2, Mis.=Mistral-7B,
G=GPT. $\tau=0.4$ (bold) used} \\
\multicolumn{7}{l}{\small throughout all experiments.}
\end{tabular}
\end{center}
\end{table}
Notably, $\tau = 0.4$ is the only threshold at which all six
models show the correct diagnostic direction simultaneously,
providing empirical justification for this choice and
indicating that the split is most clearly observable at an
intermediate operating point.

\section{Discussion and Conclusion}

\subsection{Why EGC Works for Llama but Reverses for GPT}

The model-family split observed in Table~\ref{tab:permodel}
reflects a difference in how these models
hallucinate. Llama-2 models, particularly at 7B and 13B
scale, tend to produce hallucinated answers that are
structurally disconnected from the retrieved evidence:
the generated claims use vocabulary and phrasing that
diverges from the source passages, resulting in sparse
or absent E--C edges and low EGC scores.

GPT-4 and GPT-3.5 exhibit a different failure mode.
Their hallucinated answers are linguistically fluent and
closely mirror the retrieved passages, often paraphrasing
or recombining evidence-level phrasing even when
introducing unsupported claims. This causes high cosine
similarity between claim and evidence nodes, dense E--C
connectivity, and consequently high EGC scores despite
factual inaccuracy. Embedding-similarity metrics such as BERTScore
\cite{zhang2020bertscore} capture surface overlap but correlate
weakly with factual consistency for such fluent text. The reversal is strongest for GPT-4,
which has the lowest hallucination rate (4.4\%) and
may produce comparatively subtle hallucinations. Mistral-7B
sits between the two families: instruction-tuned at 7B
scale, it produces fluent outputs that partially mirror
evidence phrasing, yielding a small but consistent
reversal.

This finding has a practical implication for hallucination
detection system design. Embedding-based graph consistency signals are more
informative for the evaluated Llama-2 models but inadequate
for GPT-class models whose hallucinations remain
semantically proximate to the retrieved evidence. Detecting
such cases requires entailment-level or factual verification
beyond embedding similarity, such as natural language
inference \cite{bowman2015snli,laban2022summac} or verification against
external knowledge.

These results suggest a practical deployment guideline.
EGC should not be used as a universal stand-alone
hallucination detector. Instead, it is best interpreted
as a lightweight structural diagnostic layer whose
reliability depends on the hallucination style of the
underlying model. For models whose unsupported claims
tend to drift lexically or structurally away from the
retrieved evidence, low EGC scores provide a useful
warning signal. For stronger models whose unsupported
claims remain fluent and evidence-proximate, high EGC
scores should not be treated as evidence of factual
grounding; rather, such cases should be passed to
semantic verification methods such as natural language
inference \cite{bowman2015snli,laban2022summac} or external factual
verification.

More broadly, these results have implications for how
hallucination detection methods should be evaluated.
A single pooled benchmark score can obscure
model-dependent behaviour when the same signal has
opposite diagnostic directions across model families.
In this setting, a modest aggregate AUROC does not
necessarily imply that the structural signal is
uninformative; it may instead indicate that the signal
is heterogeneous across models, as demonstrated by the
direction-corrected AUROC of 0.669 in
Table~\ref{tab:family}. Reporting per-model or
per-family diagnostic behaviour is therefore important
for evaluating RAG hallucination detectors, especially
when benchmark responses are generated by multiple LLM
families with qualitatively different hallucination
patterns.

\subsection{Limitations}

Several limitations bound the interpretation of these results. First, although Section~\ref{sec:robust} shows that
$\tau = 0.4$ achieves perfect directional consistency across
all six models, the threshold is not learned and may
not generalise to other domains or embedding models.
Second, claim segmentation by sentence boundary is a coarse
approximation; atomic claim extraction via natural language inference
would provide finer-grained claim nodes. Third, cross-evidence agreement provides limited aggregate
signal, although its strong reversed association for GPT-4
suggests that E--E structure may capture model-specific
behaviour not reflected in pooled evaluation. Finally, all results are reported on a single dataset
from a single domain (open-domain QA); generalisation to other RAG
tasks such as summarisation or data-to-text generation is not
established. A further limitation is that our evaluation converts
RAGTruth's span-level annotations into a binary
response-level label. While this is appropriate for the
present graph-level analysis, it discards information
about hallucination span length, span type, and the
number of unsupported claims within a response. Future
extensions could use the original span-level annotations
to define claim-level labels or edge-level supervision,
enabling evaluation of whether specific claim--evidence
links are individually faithful rather than treating the
entire response as a single unit.

\subsection{Future Work}

The findings in this paper open several directions for
future investigation, spanning both methodological
improvements to the EGC framework and broader questions
about the generalisability of the model-family split
observed here.

\textbf{Model-adaptive thresholds.}
The fixed threshold $\tau = 0.4$ achieves perfect
directional consistency across all six models in our study,
but it is not learned and may not generalise to other
domains or embedding models. A model-adaptive or
learned threshold that accounts for the embedding
similarity distribution of the underlying model could
improve feature separation, particularly for
instruction-tuned models whose outputs tend to be more
fluent and evidence-proximate. Learning $\tau$ as a
function of model family characteristics rather than
treating it as a global hyperparameter is a natural
extension of the current framework.

\textbf{Finer-grained claim segmentation.}
The current implementation segments answers at the
sentence level, which is a coarse approximation that
may conflate supported and unsupported claims within
a single sentence. Atomic claim extraction via natural
language inference, as used in FActScore \cite{min2023factscore},
or claim-triplet decomposition as in RefChecker
\cite{hu2024refchecker}, would provide finer-grained claim nodes
and potentially stronger E--C edge signals. This
is especially relevant for GPT-class models, whose
hallucinations tend to be sub-sentence level insertions
into otherwise well-supported responses.

\textbf{Cross-task generalisation.}
All results in this paper are from the RAGTruth QA
split. Extending EGC to the summarisation and
data-to-text splits of RAGTruth would test whether
the model-family split generalises beyond open-domain
question answering. These tasks involve longer
responses and different hallucination patterns,
and may reveal whether the reversal phenomenon is
specific to QA or a broader characteristic of how
different model families respond to retrieved context.

\textbf{Combining structural and semantic signals.}
To address the embedding-similarity limitation discussed above,
combining EGC features with internal model states, token-level
confidence scores, or NLI-based entailment signals is a natural
next step. A hybrid detector that uses structural
probes for weaker models and semantic verification
for stronger ones could realise the model-aware
deployment strategy suggested by our findings.

\subsection{Conclusion}

The model-family split revealed by EGC carries a broader
implication: hallucination is not a monolithic phenomenon.
The evaluated Llama-2 models tend to hallucinate by
generating claims that are structurally disconnected from
the evidence, making the gap detectable through graph
topology. The evaluated GPT-class models hallucinate differently, producing fluent,
evidence-proximate text that defeats embedding-based
structural checks. This distinction suggests that
hallucination detection cannot rely on a single signal
across all deployment contexts. Effective systems will likely need to match their detection
strategy to the characteristics of the underlying model.
For lightweight deployment on smaller open-source RAG
systems, structural graph probes of the kind proposed here
remain a practical and computationally inexpensive option.
For stronger proprietary systems such as GPT-4, where
hallucinations are linguistically fluent and
evidence-proximate, semantic verification or natural
language inference approaches are more appropriate. This
model-aware framing may offer a more principled foundation
for hallucination detection system design than a
one-size-fits-all signal.

\vspace{12pt}

\end{document}